\documentclass[11pt]{article}
\usepackage[utf8]{inputenc}
\usepackage[T1]{fontenc}
\usepackage{newtxtext}
\usepackage{newtxmath}
\usepackage{amsmath}
\usepackage{graphicx}
\usepackage{booktabs}
\usepackage{xcolor}
\usepackage[letterpaper,margin=1in]{geometry}
\usepackage[hidelinks]{hyperref}
\usepackage{natbib}
\graphicspath{{figs/}}

\newcommand{\within}{H_1^{\textnormal{within}}}
\newcommand{\cross}{H_1^{\textnormal{cross}}}

\newcommand{\va}{\textnormal{va}}
\newcommand{\mem}{\textnormal{mem}}
\newcommand{\Ceff}{C}
\newcommand{\fm}{\textnormal{FM0}}

\newif\ifanon
\anonfalse
\title{Measuring Memory and Generalization as Separable Geometric Channels:\\
The Topo$^2$ Framework}

\ifanon
\author{Anonymous Author(s)}
\else
\author{Zhanbo Zhang\thanks{Corresponding author: \texttt{553234641@qq.com}}\\[1ex]
  Ming Liu\\[1ex]
  Qing Wang\\[2ex]
  \small Anhui iFlytek Yinglian Technology Co., Ltd.}
\fi
\date{}

\begin{document}
\maketitle

\begin{abstract}
Noisy-trained deep networks simultaneously generalize on clean data and
memorize flipped labels. These are usually conflated as pressures on one
capacity. We present Topo$^2$, a measurement framework that makes them
\textbf{causally separable, measurable, and law-governed}. Persistent-homology
$H_1$ structure of the representation space separates into a within-class
manifold channel $\within$ (a function of the training stopping point) and a
cross-class channel $\cross$ (a monotone readout of memorized flipped samples).
An intervention, the \fm{} prescription (zero loss on flipped samples from
epoch 0), reaches each setting's generalization ceiling while memorizing
essentially nothing. Within the framework we establish a law set with graded
evidence: (L2) \fm{} separation prescription (9/9); (L1) $\within$ as a
training-position function (mid-rise 6/6; convergence-back CIFAR 3/3, SVHN
2/3); (L3) a ring-construction identity (definitional, not a law); and TLS
(memory--generalization topological layering): memory is \emph{causally
additive}, \emph{anchored} (silencing clean collapses the representation),
\emph{invertible} (stripping memory restores near-ceiling generalization), and
\emph{quantitatively billable} (the memorization cost law, effective slope
coefficient $\Ceff\approx0.38$ at the reference capacity: CIFAR-10 0.3801 /
SVHN 0.3806 / CIFAR-100 0.384 / VGG 0.3715, capacity-dependent in general and
traced to clean-sample feature displacement). We also publish the framework's
boundaries: a
falsification ledger of nine dead ends, and an instrument-vindication section
that excludes six families of global statistics as explanations of $\within$.
The framework turns ``memorization'' from an ill-defined capacity into a
measurable, separable, invertible topological layer.
\end{abstract}

\section{Introduction}\label{sec:intro}

\subsection{The measurement problem}\label{sec:meas-problem}

``Does the model memorize the noise?'' is a question every label-noise paper
asks, but it is usually answered by a scalar (training loss on noisy samples),
which conflates memorization with overfitting, and which cannot distinguish
\emph{where} in the representation the memorized information lives or
\emph{whether it can be removed}. The Topo$^2$ framework replaces this with a
\textbf{geometric measurement + causal intervention} toolkit.

A framework of this kind earns its keep in two ways that a single experiment
cannot: (i) the instrument must be \textbf{vindicated} --- shown to read what it
claims to read, against alternatives that could produce the same numbers by
accident (Sec.~\ref{sec:instrument}); and (ii) the framework must state
\textbf{what it does not claim}, including a record of hypotheses it actively
killed (Sec.~\ref{sec:ledger}). We take both obligations literally.

\subsection{Contributions}\label{sec:contributions}

\begin{enumerate}
\item \textbf{A measurement protocol} for the two channels ($\within$, $\cross$)
  with documented normalization obligations (pipeline-fixed $\cross$, rng-noise
  bounds, dual-scope memory readout) and an instrument-vindication section that
  excludes six families of global statistics as explanations.
\item \textbf{A causal intervention} (FM0) that produces a clean generalization
  substrate, enabling overlay (add memory), strip (remove memory), and re-anchor
  (memory with/without $G$) experiments.
\item \textbf{A law set with graded evidence} (cross-sectional $\bigstar$,
  trajectory $\lozenge$, causal $\bullet$) and a running ledger of what was
  \emph{falsified} --- we argue a framework paper should publish its dead ends as
  loudly as its laws.
\item \textbf{A determinism analysis} showing that $\within$ is a
  \emph{deterministic function of the RNG sequence} (eval cadence included as a
  training variable) --- a necessary condition for turning the trajectory law
  from observation into prediction.
\item \textbf{Honest boundary conditions}: what the framework cannot claim
  (independent prediction via $\cross$; method competitiveness; the residual
  image term in $\Ceff$; robustness of the ``break'' beyond single-seed
  decompositions).
\end{enumerate}

\section{Measurement Methods}\label{sec:method}

\subsection{Instrument vindication (why $\within$ reads what we say it reads)}\label{sec:instrument}

Before any law can rest on $\within$/$\cross$, we must rule out that the
readout is an artifact of the pipeline or a proxy for some already-known global
statistic. We do this in three independent moves.

\subsubsection{The exclusion chain: six global-statistic families falsified}

\begin{figure}[t]
\centering
\begin{minipage}[t]{0.485\textwidth}
\centering
\includegraphics[width=\linewidth]{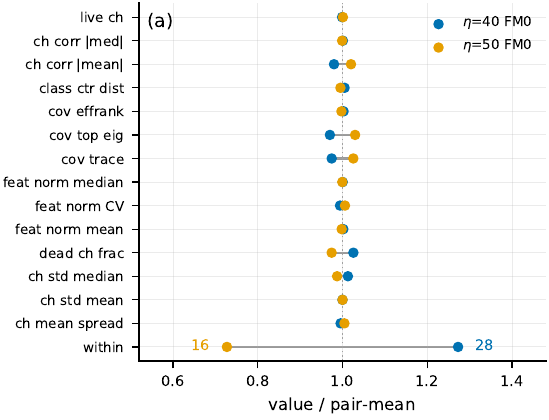}
\end{minipage}\hfill
\begin{minipage}[t]{0.485\textwidth}
\centering
\includegraphics[width=\linewidth]{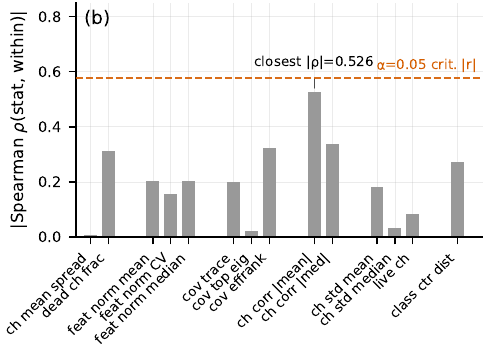}
\end{minipage}
\caption{Instrument vindication by exclusion.
\emph{(a)} The decisive pairing: $\eta$40-FM0 vs $\eta$50-FM0, same
architecture, dataset, seed and protocol. Every one of the 14 channel
statistics (all normalized to pair-mean, $x=1.0$) is near-indistinguishable
between the two checkpoints ($\leq$6\% relative deviation), and val\_acc agrees
to 0.3\% (0.966 vs 0.963) --- yet $\within$ differs by 12 (28 vs 16, the only
row drawn with numerical labels). Two models that every global statistic treats
as the same are separated by the topological readout.
\emph{(b)} All 14 statistics, grouped into six semantic families, plotted as
$|$Spearman $\rho$(stat, within)$|$ over the 12-checkpoint CONV trajectory.
None crosses the $\alpha{=}0.05$
critical value ($|r|=0.576$, dashed red); the closest statistic is annotated
once. Six families falsified.}
\label{fig:vindication}
\end{figure}

A skeptic might say $\within$ is just class-wise density, or neural collapse, or
feature-norm, or spectral shape, or memorization, or a channel-statistic
accident. We tested each family directly and each failed to reproduce or predict
$\within$ (Table~\ref{tab:exclusion}; Fig.~\ref{fig:vindication}).

\begin{table}[h]
\centering
\caption{The exclusion chain: six global-statistic families, each tested
directly and each falsified as an explanation of $\within$.}
\label{tab:exclusion}
\footnotesize
\begin{tabular}{p{2.2cm}p{3.4cm}p{1.1cm}p{5.3cm}}
\toprule
\textbf{Global-statistic family} & \textbf{Probe} & \textbf{Verdict} & \textbf{Decisive number} \\
\midrule
Class-wise density (kNN density gradient) &
  density-breach probe (task 1) & falsified &
  Breakpoint's kNN-distance CV change ($-0.32$) is same order as non-breaking
  settings ($-0.35$/$-0.54$/$-0.24$); the synthetic manual predicts a density
  spike would break $\within$, but the real break is \emph{not} driven by it \\
Neural collapse &
  density-breach probe (task 2) + NC metrics & falsified &
  Breakpoint NC1$=2.25$ $\approx$ other $\eta$40 settings (2.06--2.30); collapse
  is preserved (ETF\_cos 0.985) yet $\within$ still breaks $\to$ collapse does
  not protect uniformity here \\
Feature norm &
  feature-norm probe & falsified &
  Normalization does not restore $\within$: $\eta$40 ORIG within mean $22.3$
  $\to$ norm $24.0$ (within stays low under the norming that should remove a
  norm driver) \\
Spectral shape &
  spectral-shape probe & falsified &
  Participation ratio of class covariances is $\eta$50 ORIG 6.69 vs CONV 6.68
  (nearly identical) while $\within$ differs 23.0 vs 30.3 (a gap of 7.3); a
  spectral summary cannot carry the within difference \\
Memorization amount &
  memory probe & falsified &
  $\eta$40 lr01 (non-break) mem 0.97, $\eta$40 CONV (break) mem 1.0: both
  near-saturate, yet within differs 30 vs 22 (a gap of 8) \\
Channel statistics &
  channel-stat probe & falsified &
  14 channel statistics across 12 checkpoints: none significantly correlates
  with $\within$ \\
\bottomrule
\end{tabular}
\end{table}

The \textbf{decisive pairing} is an FM0 match: \texttt{eta40\_FM0} (within$=28$)
vs \texttt{eta50\_FM0} (within$=16$). The two checkpoints are numerically
near-indistinguishable: val\_acc agrees to 0.3\% (0.966 vs 0.963), and all 14
global feature statistics agree to
within $\leq$5.8\% relative deviation (median 0.9\%, largest:
\texttt{cov\_top\_eig\_mean} 2.04 vs 2.16). Yet $\within$ differs by 12 (28 vs
16). Two models that every global statistic --- and val\_acc --- treats as the
same are still separated by the topological readout. $\within$ carries
information no global statistic carries.

\subsubsection{Positive calibration: what $\within$ actually measures}

Reading the pipeline source and running decisive synthetic constructions (C2,
10 classes $\times$ 200 points $\times$ 512 dims, 3 seeds) pins the semantics:

\begin{itemize}
\item \textbf{$\within$ = the local-density homogeneity of class-local geometry
  along intra-class interpolation paths} (not ``class-manifold topological
  complexity''). Homogeneous construction (isotropic, locally identical)
  $\to$ high within ($32.3\pm2.6$, 3 seeds); hetero-scale construction (locally
  varying) $\to$ lowest ($13.7\pm2.4$).
\item The \textbf{synthetic semantic manual} quantifies the dominant factor:
  \textbf{density variance is the single largest driver} (sensitivity $s$:
  $0 \to 37.0$, $0.3 \to 24.3$, $0.6 \to 17.7$, $1.0 \to 3.0$ --- one factor
  alone can crush $\within$). Intrinsic dimension / anisotropy are flat
  ($\approx$26--36, no trend).
\item \textbf{$\cross$ = neighborhood label mixing}, with a monotone dose--
  response to inter-class centroid distance ($\delta{=}0 \to 157$,
  $\delta{=}24 \to 34$ floor). Crucially, \textbf{topological interleaving does
  not trigger cross} (interleaved crescent 31, orthogonal crossing 34
  $\approx$ separated-Gaussian floor 34) --- cross reads local label mixture,
  not manifold-level link/interleaving structure.
\item Pipeline robustness (real ResNet representations): the 9-cell
  \texttt{N\_NEIGHBOR} $\times$ \texttt{PCA\_DIM} core asymmetry holds with no
  sign flips (cross slope $+129\ldots+173$, within slope $-3\ldots+8$).
\end{itemize}

Two self-corrections recorded here, both tightening rather than expanding:
(i) \textbf{G1 Gaussian null} --- pure Gaussian noise through the same pipeline
returns within$=30.3\pm8.7$ (512-d) / $30.3\pm6.0$ (64-d), identical to random
and trained networks, so the specific \emph{value} ``within$\approx$30'' is a
\textbf{pipeline artifact}, not an architecture constant; the invariant is
correctly stated as \textbf{zero deviation from the random baseline}, not ``the
value 30''. (ii) \textbf{C2 semantics} --- within measures \emph{homogeneity},
not topological richness; every earlier conclusion (response asymmetry,
prediction-power asymmetry, intervention responses, baseline-relative
statements) depends only on within's \emph{response/predictive} behavior and
survives verbatim.

\subsubsection{Third-party cross-checks}

\begin{itemize}
\item \textbf{Neural collapse (NC), pre-registered} (three predictions, no
  post-hoc revision): all three confirmed.
  (1) \textbf{Collapse blind spot} --- clean training collapses class-internal
  variance (NC1$=0.04$--0.05, ETF\_cos 0.99+, kNN acc$=1.0$) while $\within$
  stays flat: collapse is a \emph{uniformity-preserving flow} invisible to
  $\within$. (2) \textbf{Memory resists collapse} --- flipped samples stay
  scattered in class space (flip\_disp positive), isolated outside class
  structure. (3) \textbf{Memory is a local phenomenon} --- NC metrics organize
  monotonically (ETF\_cos 0.4$\to$0.99) while cross is U-shaped and $\within$
  stays flat (28--38): memorized samples perturb local order without destroying
  mean-level ETF. An independent instrument confirms our readout is blind to
  what it should be blind to, and sees what it should see.
\item \textbf{ViT recalibration} --- the earlier ``ViT cross response weak''
  ($\rho{=}0.483$) was a layer-misalignment artifact: it used ViT \emph{block4}
  intermediate features while CNNs used final-layer features. Re-measured on the
  final CLS token (matching depth), cross slope recovers to $+142\ldots+156$
  (ResNet reference $+173$) --- the core asymmetry holds on ViT. Rule extracted:
  \textbf{cross-architecture comparison must align extraction to the final
  representation layer}; ViT features being more isotropic (PCA30 explains
  0.69--0.80 vs ResNet 0.97--0.99) is a real architectural difference, but it
  does not change the directional verdict.
\end{itemize}

\subsection{The $H_1$ pipeline}\label{sec:pipeline}

\texttt{cross\_class\_decomposition} computes $\within$/$\cross$ on the
test-feature manifold:
\begin{itemize}
\item \textbf{$\within$}: $H_1$ within same-class submanifolds.
\item \textbf{$\cross$}: $H_1$ of the cross-class complex (class-connecting
  loops).
\end{itemize}

\textbf{Normalization obligations (we document these as report-scope
requirements):}
\begin{itemize}
\item \textbf{cross is a pipeline-normalized quantity}: absolute value scales
  with \texttt{N\_PAIRS} $\times$ \texttt{N\_INTERP}. Any comparison must fix the
  pipeline and state it. Robustness check: varying \texttt{N\_PAIRS}/
  \texttt{N\_INTERP} $\times 4$ leaves cross\_fraction at 0.86--0.87.
\item \textbf{rng sampling noise $\pm$15--30\%} on absolute cross.
  Fine-grained cross-model contrasts with $\Delta<30\%$ need error bars; large
  effects (FM0$\leftrightarrow$CONT$\leftrightarrow$full overlay, $\geq 2\times$)
  are unaffected.
\item \texttt{N\_NEIGHBOR} / \texttt{PCA\_DIM}: robust (see Sec.~\ref{sec:instrument}).
\end{itemize}

\subsection{Memory readout: the dual-scope \texttt{mem\_noisy}}\label{sec:memnoisy}

\texttt{mem\_noisy} = fraction of flipped train samples whose prediction equals
the \emph{noisy} label, on a no-aug unshuffled loader. The ``dual scope'' pairs
it with \texttt{mem\_clean}, the fraction of clean train samples predicted at
their true label (the probe's \texttt{agree\_true} field);
\texttt{mem\_noisy} is the decisive readout and is used throughout. This
single readout is
consistent and decisive across settings: \textbf{CONT $\approx$ 0.999
(memorizes everything), FM0 $\approx$ 0.004 (memorizes nothing), K-half overlay
$\approx$ 0.50.} (We document a data pitfall: the legacy \texttt{mem\_rate}
field in some \texttt{fm0\_boundary} JSON is buggy and must not be used.)

\subsection{The intervention (FM0)}\label{sec:fm0}

Zero loss on flipped samples from epoch 0. Result: generalization at the setting
ceiling (SVHN $\approx$ 0.96, CIFAR-10 $\approx$ 0.91), memory $\approx$ 0. A
data-driven mask (FM0-hat) recovers 90--97\% without the noise distribution.
FM0 is an oracle; the FM0-hat mask is \emph{deployable} (L2) and provides the
substrate for all causal operations below.

\subsection{Causal operations on memory}\label{sec:causal}

\begin{itemize}
\item \textbf{Overlay (add):} from an FM0 checkpoint, un-freeze $K$ flipped
  samples (clean loss always on). $K$: $0 \to$ ALL interpolates FM0 $\to$ CONT.
  Must use the augmented FM0 training loader (no-aug $K{=}0$ control collapses
  --- documented pitfall).
\item \textbf{Strip (remove):} continue a CONT network clean-only; memory drains
  (mem $0.999\to0.11$), va recovers toward the FM0 ceiling.
\item \textbf{Re-anchor (memory without $G$):} train memory with clean signal
  silenced --- representation collapses (CIFAR-10: va $0.92\to0.04$). Memory
  \emph{cannot} be layered onto collapsed $G$.
\end{itemize}

\subsection{Determinism and the eval cadence (a methodological core asset)}\label{sec:determinism}

Same seed, same protocol, same mask, logically identical training code ---
\textbf{the only difference being the intermediate-eval cadence --- yields
distinct $\eta$50 300-ep $\within$ values} (full-test: 16 / 26 / 25 for cadences
25 / none / 100), while val\_acc changes by only $\sim$0.01 (0.6119 / 0.6135 /
0.6233). An earlier $N{=}2000$-subsample run reported 21 / 28 / 31--32; those
were subsampling offsets, retested to the full-test values above. A cadence
sweep (0/25/50/100/200) fills the set to $\{16, 22, 25, 26\}$ (within =
26/16/16/25/22 respectively). Mechanism, traced to a single line: PyTorch's
\nolinkurl{_BaseDataLoaderIter.__init__} draws its base seed via
\nolinkurl{torch.empty((), dtype=torch.int64).random_()}, \textbf{consuming the
global torch RNG}; each \texttt{iter(test\_loader)} during an intermediate eval
therefore perturbs the RNG, changing the subsequent train-loader shuffle, hence
the training-data order, hence the attractor.

We closed the causal chain with a \textbf{waste control}: no eval, but burning
the exact same number of RNG
draws at the same epochs ($w{=}1$ reproduces the eval's RNG footprint
position-precisely). The waste version reproduces the eval version
\textbf{exactly} (within equal, va bit-identical to 1e-4) $\to$ the eval's
causal channel is uniquely the single global RNG draw in
\texttt{iter(test\_loader)}. No larger mechanism needed.

\begin{figure}[t]
\centering
\includegraphics[width=0.72\textwidth]{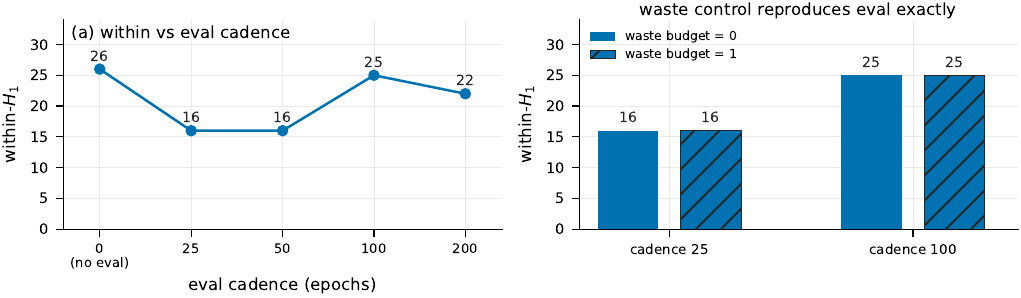}
\caption{Determinism: $\within$ as a deterministic function of the eval cadence.
\emph{(a)} Same seed, protocol, mask and training code --- only the
intermediate-eval cadence (0 / 25 / 50 / 100 / 200, equally spaced categorical
axis) differs. $\within$ takes values 26 / 16 / 16 / 25 / 22 (each labeled) while
val\_acc moves by only $\sim$0.01. \emph{(b)} The waste control ($w{=}1$: same
number of RNG draws at the same epochs, no eval) reproduces the eval condition
exactly --- within equal, va bit-identical to 1e-4 --- pinning the causal channel
to the single global RNG draw in \texttt{iter(test\_loader)}. The geometric
coordinate resolves (16--26, 38\% spread) what the scalar (2\% spread) cannot.}
\label{fig:determinism}
\end{figure}

Three consequences (Fig.~\ref{fig:determinism}):
\begin{enumerate}
\item \textbf{The training trajectory is a deterministic function of the RNG
  sequence.} Same cadence, 2 reps, bit-identical (4/4); waste reproduces the
  full state (weights + va + within); 12 independent trainings, zero deviation.
  ``Attractor'' is deterministic, not chaotic.
\item \textbf{The geometric coordinate carries more information than the
  generalization scalar.} va ranges over 0.013 (0.6105--0.6233, a 2\% relative
  spread) while within ranges over 10 (16--26, a 38\% relative spread): the ring
  coordinate resolves states the scalar cannot --- a direct, decisive,
  reproducible argument for \emph{why} a geometric measurement is needed at all.
\item \textbf{Methodological rule:} any same-seed reproducibility claim must
  declare the eval/save cadence. Intermediate checkpointing (with eval) is
  itself a training variable. All comparisons mixing checkpoints from different
  cadences need qualification.
\end{enumerate}

\section{The Law Set}\label{sec:laws}

\subsection{Evidence grading}\label{sec:grading}

$\bigstar$ cross-sectional $\cdot$ $\lozenge$ trajectory (observational) $\cdot$
$\bullet$ causal (intervention).

\subsection{L1 --- $\within$ is a training-position function ($\lozenge\bigstar$)}\label{sec:L1}

\subsubsection{The SVHN 13-point trajectory}\label{sec:L1-trajectory}

\begin{figure}[t]
\centering
\includegraphics[width=\textwidth]{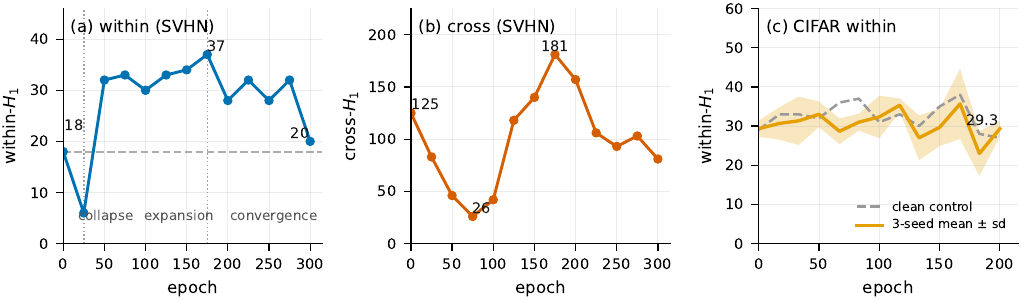}
\caption{The trajectory law. \emph{(a)} SVHN $\within$ (VGG, $\eta$40, s42,
CONV, 13 checkpoints, single axis): random-init 18 $\to$ collapse 6 $\to$
expansion plateau 30--37 (peak 37 at ep175) $\to$ convergence 20. Vertical
reference lines at ep25 / ep175 delimit the three phases
(collapse / expansion / convergence, labeled at the bottom). The dashed grey
line is the random-init baseline.
\emph{(b)} SVHN $\cross$ on the same epochs (single axis): 125 $\to$ trough 26
at ep75 $\to$ monotone rise to the ep175 peak 181, before settling to the
converged 81 --- a monotone readout of memorized flipped samples.
\emph{(c)} CIFAR reproduction (ResNet, $\eta$50): 3-seed mean $\pm$ sd (orange
band) over the clean control (grey dashed, flat). The CIFAR trajectory
\emph{partially} reproduces SVHN: converge-to-baseline is 3/3 (strong), while
the mid-rise is a weak single-point spike (Sec.~\ref{sec:L1-cifar}); the clean
control is flat, showing the convergence descent is specific to fitting
noise.}
\label{fig:trajectories}
\end{figure}

$\within$ tracks the training stopping point, not memory. VGG$\times$SVHN,
$\eta$40, s42, CONV protocol, 13 checkpoints (Table~\ref{tab:l1traj},
Fig.~\ref{fig:trajectories}):

\begin{table}[h]
\centering
\caption{SVHN 13-point trajectory (VGG, $\eta$40, s42, CONV protocol).
Three phases:
\textbf{init-collapse} (0--25: random 18 $\to$ collapse to 6, features
unexpanded) $\to$ \textbf{learning expansion} (50--175: within plateau 30--37,
val\_acc optimal at $\sim$50 ep) $\to$ \textbf{convergence collapse} (225--300:
memory saturation mem\_noisy$\to$0.99 plus an overfitting jump coincide, then
within drops to 20).}
\label{tab:l1traj}
\footnotesize
\begin{tabular}{lccccccccccccc}
\toprule
epoch & 0 & 25 & 50 & 75 & 100 & 125 & 150 & 175 & 200 & 225 & 250 & 275 & 300 \\
\midrule
within & 18 & 6 & 32 & 33 & 30 & 33 & 34 & 37 & 28 & 32 & 28 & 32 & 20 \\
mem\_noisy & .10 & .09 & .02 & .06 & .13 & .47 & .65 & .83 & .88 & .99 & .99 & 1.0 & 1.0 \\
\bottomrule
\end{tabular}
\vspace{2pt}
{\footnotesize \texttt{mem\_noisy} is the validated memory readout
(Sec.~\ref{sec:memnoisy}): the fraction of flipped train samples predicted at
their noisy label, recomputed from this run's 13 intermediate checkpoints over
the full train set (no-aug, unshuffled).}
\end{table}

\subsubsection{CIFAR reproduction, clean control, and an honest bug disclosure}\label{sec:L1-cifar}

ResNet$\times$CIFAR-10 reproduces the shape: random 29 $\to$ mid 33--37 $\to$
converged 27 (Fig.~\ref{fig:trajectories}c). Across 3 seeds the
\textbf{convergence-back-to-random-baseline is 3/3 strong} (rand $29.3\pm2.1
\to$ conv $29.3\pm2.1$, $\Delta{=}+0.0$). Two honest qualifications. First, the
CIFAR \emph{mid-rise} is weak (a single-point spike). Second, the CIFAR
\emph{convergence descent} is shallower and noisier than the SVHN 13-point
trajectory: SVHN falls from a 37 peak to a 20 endpoint (a 17-point descent on
one seed), whereas CIFAR endpoints are 27--32 across seeds (seed sd $2.1$, a
$\approx$5-point descent from a 33--37 plateau). The 3/3 fact of
converge-to-baseline survives, but the magnitude of the collapse is best read
off the SVHN single trajectory and is \emph{not} a per-seed CIFAR quantity.
Third, the SVHN convergence-back is itself \textbf{2/3, not 3/3}: across the
$\eta$50 3-seed set, s42 ($18\to21$) and s123 ($16\to23$) return toward the
random baseline, while s456 ($21\to43$, $+22$) does not --- a multi-attractor
outlier (seed sd $\approx10$) whose mechanism is open. The 6/6 count for L1 is
therefore specifically the \emph{mid-rise} (both datasets, all 3 seeds); the
convergence collapse is CIFAR 3/3 and SVHN 2/3.
The \textbf{cross trajectory is a robust memory signature across datasets}
(CIFAR dip 74--96 $\to$ converged 198--225, per-seed rise 102--142, mean
$\approx$127; SVHN trough 26 $\to$ peak 181 at ep175), far above the clean
control's \emph{converged} cross (48). The clean control's within is flat
(28--38), but its cross decays from 178 to 48 --- the contrast is
endpoint-only.

The \textbf{clean control} exists because of a bug, and we disclose it. The
first version of the CIFAR trajectory script \textbf{discarded the noisy-label
loader} and trained on clean labels --- a ``discarded-return'' bug producing
va$=0.943$ (clean ceiling) and cross$=48$ (flat). We kept that buggy run as the
clean control and fixed the script (noisy-label injection + a noise-rate assert
that passed at 0.500), giving V2: va$=0.558$ (in the CONT range), cross 225.
\textbf{The
bug-then-fixed run is itself the clean control}: it proves the collapse (the
trajectory's convergence descent) is \emph{specific to fitting noise} --- the
clean trajectory is flat. This disclosure is deliberate: the framework's
``honest record'' stance applies to its own code as much as to its hypotheses.

Noise robustness: clean vs noise within mean $|\Delta|{=}4.4 <$ seed sd 6.1 ---
statistically inseparable. L1's content, stated precisely: \textbf{the
trajectory shape is noise-robust} (noise changes the \emph{position} along it,
not its \emph{shape}); there is no numerical freezing.

\subsubsection{Protocol = stopping point}\label{sec:L1-protocol}

The ``protocol effects'' on $\within$ are duration effects. The three protocols
are not discrete categories but stops on one trajectory: ORIG 200ep $\to$ 37.5
(plateau), lr01 200ep $\to$ 30 (late plateau), CONV 300ep $\to$ 22 (past the
plateau, collapsed). CONV 200ep$\to$300ep moves within 28$\to$20 --- the last
100 epochs move it $\approx$8, matching the protocol-axis effect ($\approx$9.3).
The earlier ``protocol axis $>$ noise axis'' finding is thereby explained
mechanically: protocol changes duration, and duration is the trajectory's
independent variable.

\subsection{L2 --- FM0 separation prescription ($\bigstar$, the application anchor)}\label{sec:L2}

Epoch-0 zero loss on flipped samples reaches the setting's generalization
ceiling (SVHN 0.96+, CIFAR 0.91+) with memory $\approx$ 0. \textbf{Zero
exceptions across architectures (9/9), datasets, noise levels.} FM0-hat recovers
90--97\% and is deployable without the oracle.

\subsection{L3 --- ring-construction identity ($\lozenge$, instrumental)}\label{sec:L3}

\subsubsection{The identity}

\begin{figure}[t]
\centering
\includegraphics[width=\textwidth]{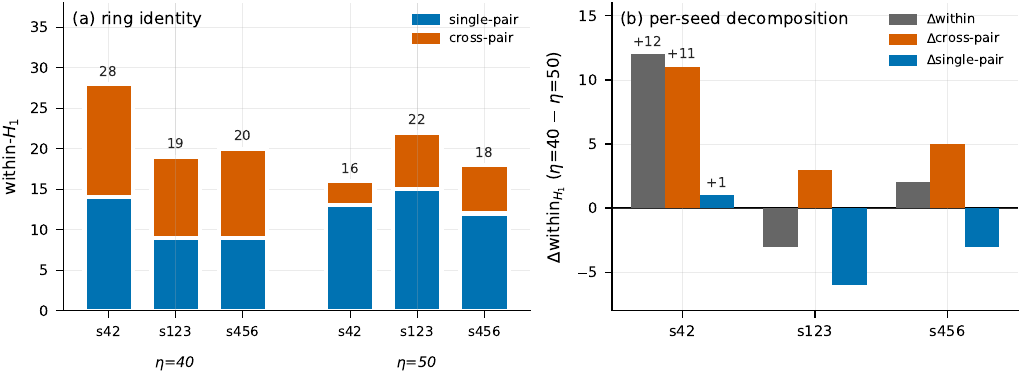}
\caption{The ring identity. \emph{(a)} For each of 3 seeds and both $\eta$40 /
$\eta$50 FM0 pairs, the stacked bar shows within $=$ single-pair rings (blue)
$+$ cross-pair same-class rings (orange), total labeled on top. The identity
within $=$ single $+$ cross holds with zero residual at all 6 checkpoints (e.g.,
$\eta$40 s42: $14+14=28$; $\eta$50 s42: $13+3=16$).
\emph{(b)} Per-seed decomposition of the FM0-pair within difference
($\eta$40 $-$ $\eta$50) into $\Delta$single $+$ $\Delta$cross. The single-seed
(s42) reading ($\Delta$within $=+12$, carried entirely by $\Delta$cross-pair
$=+11$) does not survive: s123 is single-pair dominant with the $\eta$ direction
reversed ($-6/+3$), s456 is mixed ($-3/+5$). Only the identity itself and the
FM0 val\_acc record ($\sim$0.96) survive multi-seed.}
\label{fig:ring}
\end{figure}

The $\within$ ring count decomposes as
\[
\text{within} = (\text{single-pair rings}) + (\text{cross-pair same-class rings}),
\]
holding per-model with \textbf{zero residual across 6/6 checkpoints (3 seeds)}
(Fig.~\ref{fig:ring}). Single-pair rings are the local intrinsic topology of the
10-interpolation-point graph within a position; cross-pair same-class rings
connect local graphs at different positions of the same class.

\subsubsection{Honest positioning}

This identity is \textbf{definitional composition}, not a law: it is guaranteed
by how within is computed, so its value is as a \textbf{pipeline consistency
check and a diagnostic decomposer} (what portion of within is local vs
position-linking). We state this explicitly rather than dressing a tool up as a
discovery.

\subsubsection{A multi-seed falsification of the tempting mechanism}

The single-seed (s42) FM0-pair decomposition looked mechanistic: ``the
$\eta$40-vs-$\eta$50 within difference ($\Delta{=}+12$) is \emph{entirely}
carried by cross-pair isolation ($\Delta{=}+11$), single-pair rings almost
unchanged ($\Delta{=}+1$)''. Multi-seed killed it: s123 has single-pair
dominance with the $\eta$ direction \emph{reversed} ($-6/+3$), s456 is mixed
($-3/+5$); the FM0-pair within difference itself is not robust across seeds
(s42 $+12$ / s123 $-3$ / s456 $+2$). \textbf{The ``break = cross-pair
isolation'' mechanism is an s42 single-seed decomposition; it does not
survive.} What survives the multi-seed test is only the identity itself
(definitional) and the FM0 val\_acc record ($\sim$0.96 across seeds). Recorded
in the falsification ledger (Sec.~\ref{sec:ledger}).

\subsection{TLS --- memory--generalization topological layering ($\bullet$)}\label{sec:TLS}

The four properties are listed in measured-first order: the three directly
observed properties precede the fitted cost law.
\begin{enumerate}
\item \textbf{(i) Memory is causally additive.} Overlaying memory onto the FM0
  substrate scales $\cross$ monotonically (3/3 seeds): 90$\to$221 /
  91$\to$254 / 111$\to$219. At saturation, reaches CONT-level cross.
\item \textbf{(iii) Memory anchors to $G$.} Silencing clean collapses memory
  (CIFAR-10: va $0.92\to0.04$); the FM0 mask works because $G$ stays in the room.
\item \textbf{(iv) within is secondary.} $\within$ responds to memory within
  seed noise (its response is a training-duration artifact).
\item \textbf{(ii) Memory is quantitatively billable.} The memorization cost law
  (companion paper, all numbers verified):
  \begin{equation}\label{eq:costlaw}
  \va = \va_{\fm} - \Ceff\cdot\frac{\eta}{1-\eta}\cdot\mem, \qquad
  \Ceff\approx0.38.
  \end{equation}
  $\Ceff$ at the $\sim$11M reference capacity, across 4 settings (all 3-seed
  where stated): CIFAR-10 0.3801 ($R^2{=}0.994$) / SVHN 0.3806
  ($R^2{=}0.996$) / CIFAR-100 0.384 / VGG 0.3715. Class-count independence at
  fixed capacity is verified by a decisive experiment
  (Sec.~\ref{sec:C-mechanism});
  $\eta/(1-\eta)$ beats linear / power-law / $\eta^2$ forms. $\Ceff$ is a
  \emph{capacity-dependent effective coefficient}, not a universal constant:
  a CIFAR-10 width sweep (3-seed mean) gives 0.471$\to$0.342
  (w0.5$\to$w2.0), and the four settings agree near 0.38 because they share
  the reference capacity. Its mechanism traces to a normalized clean-sample
  displacement cost ($\Ceff\approx0.5\,d_{\mathrm{clean}}$, $R^2{=}0.92$
  through origin; Sec.~\ref{sec:C-mechanism}). A $\eta$20
  low-noise extrapolation passes on SVHN ($k{=}0.087$, ratio 0.92; 5-point
  through-origin $\Ceff{=}0.3803$) and deviates on CIFAR-10 ($k{=}0.125$, ratio 1.31;
  5-point $\Ceff{=}0.3820$) --- the law's cleanest domain is the mid-noise range
  $\eta$40--60; at low noise the $\eta/(1-\eta)$ form is a first-order
  approximation with dataset-dependent curvature (see the companion paper's
  low-noise extrapolation section). Memory is an \textbf{invertible layer}:
  stripping CONT restores
  near-ceiling va.
\end{enumerate}

\begin{figure}[t]
\centering
\includegraphics[width=\textwidth]{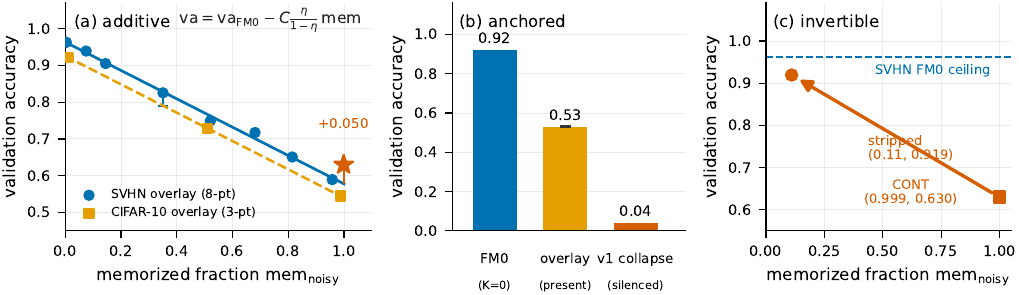}
\caption{TLS properties. \emph{(a) Additive / billable.} SVHN overlay: val\_acc
vs memorized fraction, 8 points (FM0 $\to$ K $= 2500\ldots30000 \to$ ALL), with
the cost-law fit $\va = \va_{\fm} - \Ceff\,\eta/(1-\eta)\,\mem$ drawn through
them (slope $-\Ceff\approx-0.386$ on SVHN; CIFAR 3-point overlay in orange).
K$=$12500 carries a real 3-seed error bar. Co-training (CONT, $\star$) sits
\emph{above} the overlay curve at matched memory load ($+0.050$, SVHN resnet).
The SVHN ALL endpoint shown is the ep100 checkpoint of the K$=$ALL run (mem
0.958, va 0.589); the same run's later ep160 checkpoint (mem 0.981, va 0.585)
saturates slightly higher but is not shown here.
\emph{(b) Anchored.} CIFAR three states: FM0 (K$=$0) at 0.92, full overlay with
clean present at 0.53$\pm$0.004 (3-seed), and v1 collapse with clean silenced at
0.04 --- memory cannot be layered onto collapsed $G$.
\emph{(c) Invertible.} Stripping a CONT network (clean-only continuation) drains
memory 0.999$\to$0.11 and recovers va 0.630$\to$0.919, approaching the SVHN FM0
ceiling (dashed).}
\label{fig:tls}
\end{figure}

Fig.~\ref{fig:tls} shows the three measurable properties: additivity (a),
anchoring (b), and invertibility (c).

\subsection{Co-evolution advantage (setting-specific)}\label{sec:coev}

At matched memory load, co-training (CONT) exceeds post-hoc overlay: SVHN
resnet $+0.050$ (structural), CIFAR $+0.006$, VGG $+0.013$. Setting-specific
(SVHN resnet); determining factor open.

\section{Empirical Panorama (key tables)}\label{sec:panorama}

Throughout, $\mem$ is the memorized fraction \texttt{mem\_noisy}
(Sec.~\ref{sec:memnoisy}), $\eta$ the noise rate, and ``4 settings'' means
CIFAR-10 / SVHN / CIFAR-100 / VGG$\times$SVHN with the resnet-family
architecture fixed where not stated. Per-claim statistical strength is stated
in Sec.~\ref{sec:stats}.

\begin{table}[h]
\centering
\caption{The law set at a glance. Evidence: $\bigstar$ cross-sectional,
$\lozenge$ trajectory, $\bullet$ causal.}
\label{tab:panorama}
\footnotesize
\begin{tabular}{lp{1.5cm}p{6.8cm}}
\toprule
\textbf{Law} & \textbf{Evidence} & \textbf{Key numbers} \\
\midrule
L1 within position & $\lozenge\bigstar$ 6/6 &
  SVHN 13-pt 18$\to$37$\to$20; CIFAR 29$\to$33--37$\to$27; mid-rise 6/6; \\
  & & CIFAR conv$=$rand 3/3, SVHN conv$=$rand 2/3 \\
L2 FM0 prescription & $\bigstar$ 9/9 &
  SVHN 0.96+ / CIFAR 0.91+; FM0-hat 90--97\% \\
TLS(i) additive & $\bullet$ 3/3 &
  cross 90$\to$221, 91$\to$254, 111$\to$219 \\
TLS(iii) anchored & $\bullet$ &
  va 0.92$\to$0.04 (clean silenced) \\
TLS(iv) within secondary & $\bullet$ &
  $\Delta$within within seed sd \\
TLS(ii) cost law & $\bullet$ 4 settings &
  $\Ceff$: 0.3801 / 0.3806 / 0.384 / 0.3715\textsuperscript{$\dagger$} \\
Inverse (strip) & $\bullet$ &
  va 0.63$\to$0.92; mem 0.999$\to$0.11 \\
Co-evolution & $\bullet$ &
  SVHN resnet $+0.050$ / CIFAR $+0.006$ / VGG $+0.013$ \\
Determinism & $\lozenge$ &
  within $\in\{16,22,25,26\}$ same seed, cadence-only; waste $w{=}1$ bit-identical \\
\bottomrule
\end{tabular}
\vspace{2pt}
{\footnotesize $\dagger$Effective slope coefficient $\Ceff$ of the memorization
cost law (Eq.~\ref{eq:costlaw}), one value per setting at the $\sim$11M
reference capacity, all 3-seed where stated: CIFAR-10 0.3801 ($R^2{=}0.994$),
SVHN 0.3806 ($R^2{=}0.996$), CIFAR-100 0.384, VGG (VGG$\times$SVHN, cross-seed
CV 7.7\%) 0.3715. Caliber note: 0.3715 is the VGG value and is the lowest of the
four, matching the VGG cell's larger cross-seed variance
(Sec.~\ref{sec:stats}); all four sit within 3\% of 0.38 at the shared reference
capacity. $\Ceff$ is capacity-dependent in general
(Sec.~\ref{sec:C-mechanism}). Full derivation in the companion paper.}
\end{table}

\section{What the Framework Does NOT Claim}\label{sec:nonclaims}

\subsection{$\cross$ is not an independent generalization predictor}\label{sec:cross-nonpred}

Partial correlations corr($\cross$, va $\mid$ mem), $n{=}37$: CIFAR $-0.45\ldots
-0.56$ ($\Delta R^2$ $+0.046$), SVHN $+0.08$ ($\Delta R^2$ $+0.000$). Weak and
setting-dependent $\to$ we call $\cross$ a \emph{geometric signature of memory},
not a predictor. The TLS(i) causal scaling is a different claim (monotone under
intervention) and is unaffected.

\subsection{FM0 is not a SOTA method}\label{sec:fm0-nonsota}

CIFAR-10 $\eta$40: DivideMix 0.94--0.95 $>$ FM0 oracle 0.92 $>$ ELR+ 0.91 $>$
ELR 0.89 $>$ FM0-hat 0.85--0.87 $>$ Co-teaching 0.80--0.84. FM0's value:
mechanistic ceiling + unified language.

\subsection{$\Ceff$ is a reference-capacity value, not a universal constant}\label{sec:C-mechanism}

The law's coefficient is capacity-dependent: a CIFAR-10 width sweep (3-seed
mean) gives $\Ceff$ 0.471$\to$0.342 (w0.5$\to$w2.0). The four settings agree
near 0.38 because they share a $\sim$11M reference capacity, not because
$\Ceff$ is universal. Its mechanism is traced (companion paper): at fixed image
family $\Ceff\approx0.5\,d_{\mathrm{clean}}$ (through origin, $R^2{=}0.92$),
where $d_{\mathrm{clean}}$ is the normalized clean-sample feature displacement
during overlay; a head/backbone swap shows the cost is 100\% feature-mediated.
Class-count independence at fixed capacity is verified by \emph{retraining} on
10 superclass labels (a coarse merge of the same CIFAR-100 images):
$\Ceff{=}0.370$ vs the fine
$\Ceff{=}0.369$ ($\eta$50 / $K{=}12500$ caliber). At that caliber the residual
image term is $\approx\pm0.02$ (CIFAR-10 $\approx0.40$ vs CIFAR-100
$\approx0.37$), while the law's $\eta$-averaged values (0.3801 / 0.384) agree
to $\approx$1\%.

\subsection{Statistical strength is graded, not uniform}\label{sec:stats}

3-seed cells: CIFAR-10 $\eta$30/40/50/60, SVHN $\eta$40, CIFAR-100 $\eta$30/$\eta$50,
VGG $\eta$50. Single-seed cells: SVHN $\eta$30/$\eta$50/$\eta$60 ($\eta$50
corroborated by an 8-point full-curve fit), VGG $\eta$30/$\eta$60, CIFAR-10
$\eta$20, SVHN $\eta$20. We state strength per number; cross-seed variance on
VGG $\eta$50 (CV 7.7\%) is 2--4$\times$ the resnet values.

\subsection{Falsification ledger}\label{sec:ledger}

A framework paper should publish its dead ends as loudly as its laws. Each
entry: hypothesis / experiment that killed it / lesson.

\begin{enumerate}
\item \textbf{Selective forgetting} --- \emph{Hypothesis}: CONT networks
  ``choose'' to forget some noise (earlier estimate: $\sim$49\% remembered).
  \emph{Killer}: re-measured with the pred$=$noisy readout on corrected code,
  CONT memorizes \emph{everything} (\texttt{mem\_noisy}$\approx$0.999, 9/9
  settings-seed); the 0.49 was a legacy \texttt{mem\_rate} field bug.
  \emph{Lesson}: memory-load claims must use the validated readout; a single
  data-source bug can misattribute the most central fact.
\item \textbf{Ring overload} --- \emph{Hypothesis}: CONT underperforms at
  $K{=}$ALL overlay because it ``overloads'' its rings. \emph{Killer}: at
  matched memory load (CONT 0.999 vs $K{=}$ALL 0.98) CONT \emph{exceeds}
  overlay; no load difference exists. \emph{Lesson}: matched-load comparisons
  are mandatory before structural explanations.
\item \textbf{Crossing point as law (``controlled overlay $>$ CONT'')} ---
  \emph{Hypothesis}: $K$-half overlay's va edge over CONT is a law of
  ``controlled memory''. \emph{Killer}: three confounds not separated (oracle
  mask knows the noise distribution + half the memory + co-training path)
  $\to$ downgraded to a \emph{sampling point} on the tradeoff curve.
  \emph{Lesson}: attribution confusion masquerades as discovery; the curve is
  the claim, single points are not.
\item \textbf{Concentration mechanism} --- \emph{Hypothesis}: $\Ceff$'s slope
  constant should shift on non-10-class data because noise:clean concentration
  differs. \emph{Killer}: the law-level comparison gives $\Ceff{=}0.3801$
  (CIFAR-10) vs $\Ceff{=}0.384$ (CIFAR-100), $\approx$1\%; the same CIFAR-100
  images retrained on 10 superclass labels give $\Ceff{=}0.370$ vs the fine
  $\Ceff{=}0.369$ ($\eta$50 / $K{=}12500$ caliber) --- \emph{at fixed capacity},
  how flips distribute across classes does not enter $\Ceff$.
  \emph{Lesson}: a precise quantitative prediction, when wrong, sharpens the
  law more than when it is right --- and the independence holds only at fixed
  capacity: $\Ceff$ \emph{does} depend on capacity
  (Sec.~\ref{sec:C-mechanism}).
\item \textbf{G3 generalization prediction (independent predictor)} ---
  \emph{Hypothesis}: $\cross$ independently predicts generalization gap (initial
  $+0.43$ cross-config correlation). \emph{Killer}: the within-epoch axis runs
  \emph{opposite} (organization artifact); cross-config vs within-epoch axes are
  not interchangeable; partial correlations with mem are weak and
  setting-dependent (SVHN $+0.08$, $\Delta R^2$ $+0.000$). \emph{Lesson}: a
  correlate on one gradient axis is not a predictor; control the confounding
  variable (mem) or the claim dies on the next axis.
\item \textbf{Within invariant ``value$=$30'' (G1)} --- \emph{Hypothesis}:
  within$\approx$30 is an architecture-intrinsic constant. \emph{Killer}: pure
  Gaussian point clouds through the same pipeline return $30.3\pm8.7$ --- the
  value is a pipeline artifact. \emph{Lesson}: null baselines are mandatory; the
  corrected claim (zero deviation from the random baseline) is stronger than the
  dead one.
\item \textbf{Six global-statistic mechanisms} --- \emph{Hypothesis}: $\within$
  is class density / neural collapse / feature norm / spectrum / memory /
  channel statistics. \emph{Killer}: all six falsified; decisive FM0 pairing has
  val\_acc equal to 0.3\% and all 14 channel statistics within $\leq$5.8\%
  relative deviation while within differs by 12. \emph{Lesson}: instrument
  vindication by exclusion is the difference between a measurement and a proxy.
\item \textbf{Density-gradient hypothesis for the break} --- \emph{Hypothesis}:
  the $\eta$40 break is caused by intra-class density non-uniformity (the
  synthetic manual's dominant within-killer). \emph{Killer}: the breakpoint's
  density change is same-order as non-breaking settings; the break is ``uniform
  + collapsed but within low'' --- an exclusion combination the density story
  cannot produce. \emph{Lesson}: a synthetic sensitivity factor need not be the
  real-world cause; negative results of a semantic manual are data too.
\item \textbf{``Memory saturation $\to$ break'' causal direction
  (FREEZE\_MEM)} --- \emph{Hypothesis}: fitting noise causes the break, so
  freezing memory should prevent it. \emph{Killer}: freezing memory gradients at
  ep225 \emph{deepens} the break (within 18) while setting a then-record
  (va 0.879); freezing clean collapses generalization (va 0.457). The
  break is a product of \emph{escaping} noise fitting toward pure
  generalization, not of fitting noise. \emph{Lesson}: causal direction is
  decided by intervention, not by temporal coincidence; a ``negative'' result
  reversed the mechanism.
\end{enumerate}

Also recorded (Sec.~\ref{sec:L3}): the ``break = cross-pair isolation''
decomposition is an s42 single-seed finding that multi-seed falsification
removed from the law set.

\section{Related Work}\label{sec:related}

\begin{itemize}
\item \textbf{Label-noise learning} (DivideMix \citep{li2020dividemix}, ELR
  \citep{liu2020early}, Co-teaching \citep{han2018coteaching}, and
  predecessors): method-oriented; FM0 provides a mechanistic ceiling and a
  common measurement language. See honest SOTA positioning in
  Sec.~\ref{sec:fm0-nonsota}.
\item \textbf{Memorization theory} (Zhang et al.\ 2017 \citep{zhang2017understanding}
  on why nets can memorize; Arpit et al.\ 2017 \citep{arpit2017closer} on when
  they do; Feldman \citep{feldman2020does} on \emph{whether} learning requires
  memorization of the long tail; Feldman \& Zhang \citep{feldman2020what} on
  \emph{which} samples and influence): complementary to our \emph{where} +
  \emph{cost}.
\item \textbf{Topological data analysis in deep learning} (Naitzat et al.\
  \citep{naitzat2020topology}; surveys \citep{ballester2023topological,
  pun2022persistent}): persistent homology as description; we add causal
  manipulation (overlay/strip) and a law-governed measurement protocol.
\item \textbf{Generalization theory}: bounds on capacity; orthogonal to a
  \emph{billable cost of memorization}.
\end{itemize}

\section{Discussion and Outlook}\label{sec:discussion}

\begin{itemize}
\item \textbf{What the framework buys}: a common measurement language (two
  channels, \texttt{mem\_noisy}, FM0 substrate, overlay/strip operators) that
  converts ``does the model memorize'' from a scalar heuristic into a causally
  testable geometric quantity. The determinism result (Sec.~\ref{sec:determinism})
  makes $\within$ a deterministic function of the RNG sequence --- a necessary
  condition for predictability, and what elevates the trajectory from
  observation to law.
\item \textbf{What remains open}: (i) the residual image term in $\Ceff$
  ($\approx\pm0.02$ at the $\eta$50 / $K{=}12500$ caliber, beyond the
  feature-displacement route; Sec.~\ref{sec:C-mechanism});
  (ii) determining factor of the co-evolution advantage (SVHN resnet);
  (iii) breadth --- more architectures/datasets/noise schedules; (iv) the
  \emph{dynamics} of within --- why convergence reshapes the local-neighborhood
  graph (the reshuffle is quantified but not mechanistically explained);
  (v) the ``chaos region'' ($n{=}3$ cannot distinguish
  deterministic-effect-plus-s42-anomaly from true chaos).
\item \textbf{Methodological stance}: publish falsified hypotheses (selective
  forgetting, ring overload, the crossing point as law, the concentration
  mechanism, G3 independent prediction, the value-30 invariant, the six global
  statistics, the density-gradient mechanism, the memory-saturation$\to$break
  direction) alongside confirmed laws. Dead ends delimit the law's boundary.
\end{itemize}

\section{Conclusion}\label{sec:conclusion}

Topo$^2$ makes memorization and generalization \textbf{measurable, separable,
and causally manipulable} in the representation geometry. The framework
provides: an instrument vindicated against six global-statistic alternatives
and two self-corrections (G1, C2); a validated measurement protocol; an
intervention (FM0) that cleanly separates the two channels; a law set with
graded evidence and a falsification ledger of nine dead ends; and a determinism
analysis that makes the trajectory law deterministic (seed-conditional), a
necessary condition for predictability. Its flagship quantitative
result --- the memorization cost law, with effective coefficient
$\Ceff\approx0.38$ at the reference capacity and capacity-dependence in
general --- is developed in the companion paper.

\bibliographystyle{plainnat}
\bibliography{refs}

\end{document}